\documentclass[10pt,twocolumn,letterpaper]{article}

\usepackage{cvpr}              

\usepackage[table,dvipsnames]{xcolor}

\usepackage[numbers]{natbib}
\usepackage{pifont}
\usepackage{array}\definecolor{good}{RGB}{199,233,199}    
\definecolor{bad}{RGB}{255,199,199}     
\definecolor{partial}{RGB}{255,245,199} 
\newcommand{\cmark}{\textcolor{ForestGreen}{\ding{51}}} 
\newcommand{\xmark}{\textcolor{red}{\ding{55}}} 
\newcommand{\pmark}{\textcolor{orange}{(\ding{51})}}
\newcommand{\goodcell}{\cellcolor{good}}
\newcommand{\badcell}{\cellcolor{bad}}
\newcommand{\partcell}{\cellcolor{partial}}

\newcommand{\PAR}[1]{\vspace{-0.2eM}\vskip4pt \noindent{\bf #1}}

\newcolumntype{B}[1]{>{\centering\arraybackslash}b{#1}} 

\usepackage{amsmath}
\usepackage{algorithm}
\usepackage{algpseudocode}
\usepackage{graphicx}
\usepackage{booktabs}   
\usepackage{caption}    
\usepackage{tabularx}    
\usepackage{longtable}  
\usepackage{pdflscape}
\usepackage{multirow}
\usepackage{makecell}

\definecolor{cvprblue}{rgb}{0.21,0.49,0.74}
\usepackage[pagebackref,breaklinks,colorlinks,citecolor=cvprblue]{hyperref}
\definecolor{verylightgray}{gray}{0.85}
\def\paperID{00033} 
\def\confName{3DV\xspace}
\def\confYear{2026\xspace}

\title{UniLiPs: Unified LiDAR Pseudo-Labeling with\\  Geometry-Grounded Dynamic Scene Decomposition}

\author{Filippo Ghilotti$^1$\qquad Samuel Brucker$^1$\qquad Nahku Saidy$^{1}$\vspace{0.5pt}
\and
Matteo Matteucci$^{2}$\qquad Mario Bijelic$^{1,3}$\qquad  Felix Heide$^{1,3}$\vspace{0.5pt}
\and
\centerline{\small{$^1$TORC Robotics} \quad \small{$^2$Politecnico of Milan} \quad \small{$^3$Princeton University}}\\[0.1em]
\small{\url{https://light.princeton.edu/unilips}}}

\begin{document}
\newcolumntype{Y}{>{\centering\arraybackslash}X}
\maketitle
\begin{abstract}
\noindent
Unlabeled LiDAR logs, in autonomous driving applications, are inherently a gold mine of dense $3\text{D}$ geometry hiding in plain sight - yet they are almost useless without  human labels, highlighting a dominant cost barrier for autonomous-perception research.
In this work we tackle this bottleneck by leveraging temporal–geometric consistency across LiDAR sweeps to lift and fuse cues from text and $2\text{D}$ vision foundation models directly into $3\text{D}$, without any manual input.
We introduce an unsupervised multi-modal pseudo-labeling method relying on strong geometric priors learned from temporally accumulated LiDAR maps, alongside with a novel iterative update rule that enforces joint geometric-semantic consistency, and vice-versa detecting moving objects from inconsistencies. 
Our method simultaneously produces $3\text{D}$ semantic labels, $3\text{D}$ bounding boxes, and dense LiDAR scans, demonstrating robust generalization across three datasets. We experimentally validate that our method compares favorably to existing semantic segmentation and object detection pseudo-labeling methods, which often require additional manual supervision. We confirm that even a small fraction of our geometrically consistent, densified LiDAR improves depth prediction by $51.5\%$ and $22.0\%$ MAE in the $80$–$150$ and $150$–$250$ meters range, respectively.
\end{abstract}
\section{Introduction}
Large-scale annotated datasets and increased computing power have enabled the succes of learned vision methods. Datasets like ImageNet\cite{deng2009imagenet}, PASCAL VOC\cite{pascal-voc-2010}, MSCOCO\cite{lin2015microsoftcococommonobjects}, Cityscapes\cite{7780719}, and ADE20K\cite{8100027} have driven advances in classification, detection, and segmentation. In autonomous driving, annotating large-scale data, especially $3D$ LiDAR scans, is challenging and costly due to the need for precise multi-modal alignment. Multimodal benchmarks such as KITTI\cite{geiger2012we}, nuScenes\cite{caesar2020nuscenes}, the Waymo Open Dataset\cite{sun2020scalability}, and Argoverse\cite{8953693} reflect this effort.
\begin{figure}[t]
    \centering
       \includegraphics[
            trim=0.3cm 0cm 0.3cm 0.2cm,  
            clip,
            width=\linewidth]{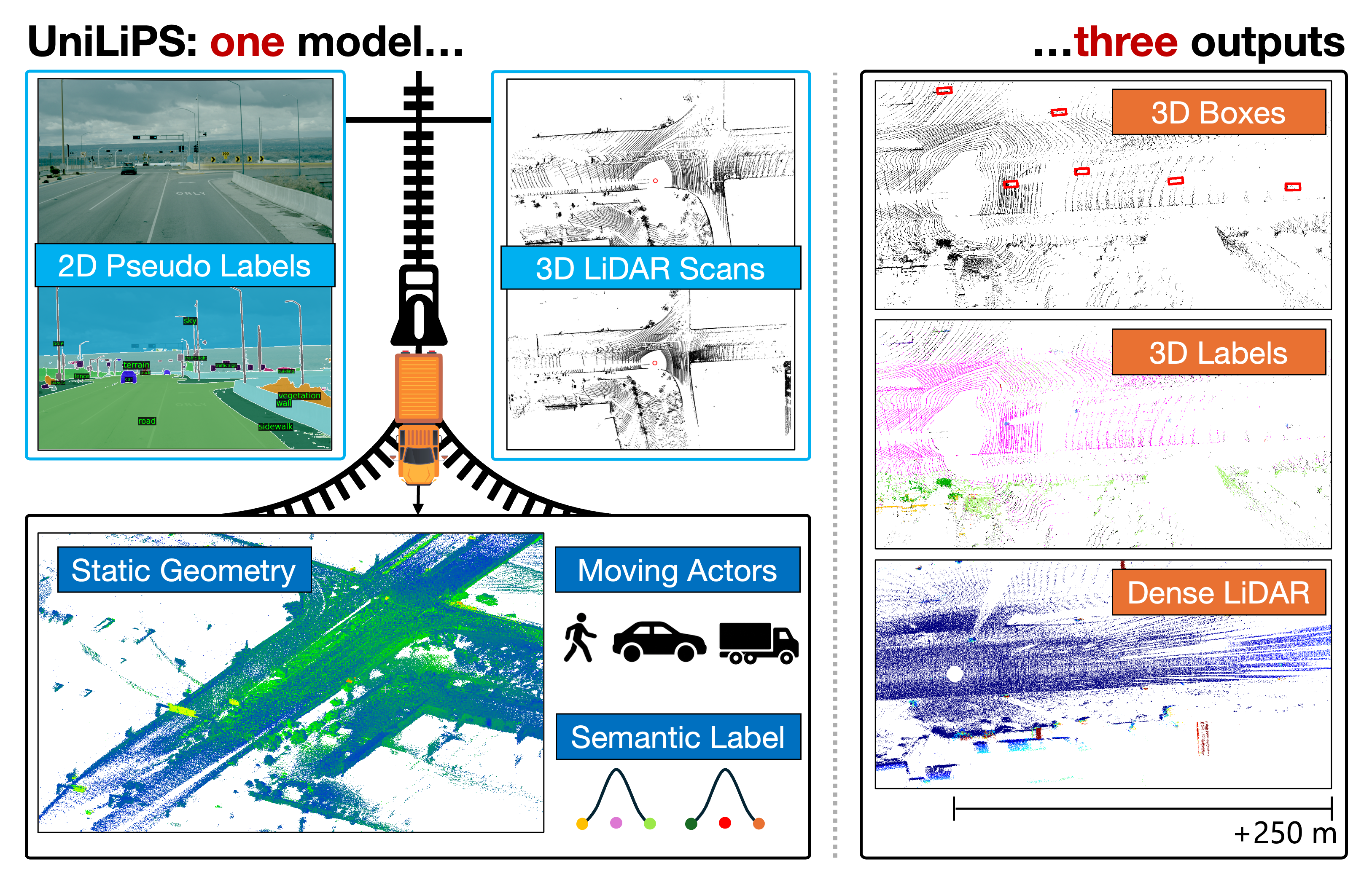}\vspace{-1mm}
    \caption{\textbf{Unified 3D Labeling.} Given a single driving trajectory, UniLiPs fuse consecutive LiDAR scans with our engine’s 2D pseudo-labels to build a coherent 3D map. Within this consistent geometry, moving actors and semantic labels are optimized to \emph{jointly} generate refined, temporally consistent 3D bounding boxes, semantic labels, and occlusion-aware, densified point clouds.
    }
    \label{fig:teaser}
    \vspace{-5mm}
\end{figure}
To tackle the annotation challenges for large datasets, a body of work explores automatic labeling, using pre-aligned $3D$ models to incorporate geometric and semantic constraints into the annotation pipeline, effectively reducing ambiguity and enhancing consistency across labels \cite{CAD-Estate, 6836101, xiang2016objectnet3D, chabot2017deepmantacoarsetofinemanytask, zakharov2020autolabeling3Dobjectsdifferentiable, li2024segmentliftfitautomatic}. Methods relying on synthetic data can generate fully annotated video sequences, providing detailed $2D$ and $3D$ multi-object tracking information, along with pixel-level labels for categories, instances, flow, and depth \cite{cabon2020virtualkitti2,dosovitskiy2017carlaopenurbandriving, 7780721}. Other efforts aim to minimize annotation workload through offline perception \cite{nuplan}, and semi-supervised approaches \cite{mustikovela2021self, zhang_depth_contrast, timoneda2024multimodalnerfselfsupervisionlidar, 10802675, nunes2023cvpr, SLidR, Baur2024ECCV, packnet-semguided, packnet} leverage unlabeled data, although depending on specialized architectures to handle partial ground-truth labels. 
Specifically, we note that these existing annotation methods typically require separate methods for each task — be it depth estimation, object detection, or semantic segmentation — and often rely on manually generated or pseudo labels that are hard to reproduce.
In contrast, we rethink the annotation process in an unsupervised, \textit{unified} $3D$ labeling framework, as presented in Figure~\ref{fig:teaser}, that concurrently tackles all these tasks by leveraging a consistent SLAM-based $3D$ map as a comprehensive semantic-geometric representation, ensuring frame invariance and enhanced reproducibility across modalities, to generate labels for tasks such as $3D$ bounding box detection, semantic segmentation, and depth estimation, with minimal parameters tuning. Our approach enriches a $3D$ map with semantic, geometric, and probabilistic information, and exploits sensor fusion and geometric consistency to automatically separate the static scene from dynamic objects. We cast the problem as a novel \textit{Iterative Update Weighted Function} to distinguish moving objects — which break the static world assumption and are refined into trajectory-aware bounding boxes — from static regions, which are then converted into densified LiDAR-like scans via \textit{Adaptive Spherical Occlusion Culling} and enhanced with rich semantic details.
We evaluate the method against held-out manual labels and training state-of-the-art networks with our pseudo-labels, on semantic segmentation, object detection and depth estimation.\\
We make the following contributions:
\begin{itemize}
    \item We introduce a novel method to obtain \textit{jointly} pseudo-labels for 3 different downstream tasks (semantic segmentation, object detection and depth estimation), at scale, with no manual annotations needed and not tied to any specific dataset or sensor suite.
    \item We devise a method for static and dynamic object  separation, exploiting points and labels temporal accumulation and an \textit{Iterative Update Weighted Function}.
    \item We find that our semantic labels and bounding boxes achieve \textit{SOTA} performances compared to standalone pseudo labeling methods and confirm they can grant \emph{close to Oracle} performance on three different datasets. 
    \item For depth estimation, we devise how a lightweight fine-tuning on a subset of our consistent pseudo ground-truth achieves \emph{improvements of 51.5\% in MAE between 80 and 150 meters and 22.0\% between 150 and 250 meters.} 
\end{itemize}

\section{Related Work}
\begin{table}[ht!]
\centering
\setlength{\tabcolsep}{4.2pt}
\renewcommand{\arraystretch}{1.15}
\begin{tabular}{lcccc}
\toprule
\makecell[tl]{\\ \\ \textbf{Outputs PL}}
 & \makecell[tc]{{Det.\ From}\\{Motion}\\\cite{Semoli, Baur2024ECCV, lin2024icp}
 }
 & \makecell[tc]{{Pseudo}\\{Seg.}\\\cite{gebraad2025leapconsistentmultidomain3d,kong2023lasermix}}
 & \makecell[tc]{{Depth}\\{Pred.}\\\cite{uhrig2017sparsityinvariantcnns, depth_anything_v2}}
 & \textbf{Ours}\\
\midrule
Bounding Boxes   & \goodcell\cmark & \badcell\xmark & \badcell\xmark & \goodcell\cmark\\
Semantic Labels  & \badcell\xmark & \goodcell\cmark & \badcell\xmark & \goodcell\cmark\\
Dense Depth & \badcell\xmark & \badcell\xmark & \goodcell\cmark & \goodcell\cmark\\
Moving Objects & \goodcell\cmark & \partcell\pmark  & \badcell\xmark & \goodcell\cmark\\
\midrule
\multicolumn{5}{l}{\textbf{Requirements and Specifications}}\\
\midrule
Long-Range & \badcell\xmark & \partcell\pmark & \goodcell\cmark & \goodcell\cmark\\
Dataset Invariant & \partcell\pmark & \badcell\xmark & \badcell\xmark & \goodcell\cmark\\
Time Consistent & \partcell\pmark & \partcell\pmark & \badcell\xmark & \goodcell\cmark\\
Unsupervised  & \partcell\pmark & \partcell\pmark & \badcell\xmark & \goodcell\cmark\\
\bottomrule
\end{tabular}
\vspace{-3mm}
\caption{\textbf{Unified Labeling. }Our approach jointly generates (\cmark ) all Pseudo Labels (PL) types, at long range, without any ground-truth supervision. By contrast, state-of-the-art methods often rely on ground-truth data, only partially (\pmark) satisfy consistency and invariance requirements and not deliver (\xmark) all the outputs.}
\label{tab:checkmarks}
\vspace{-3mm}
\end{table}
\PAR{Pseudo Depth. }High-density LiDAR depth maps are traditionally produced using LiDAR Inertial Odometry algorithms \cite{legoloam2018shan, liosam2020shan, Dong2024LiDARIO, chen2022dlio, 9718203, uhrig2017sparsityinvariantcnns} that aggregate information across multiple frames. Conversely image-based depth foundation models \cite{depth_anything_v1, depth_anything_v2, Bochkovskii2024:arxiv, https://doi.org/10.48550/arxiv.2302.12288, Ranftl2020, Ranftl2021, birkl2023midas, piccinelli2024unidepth, piccinelli2025unidepthv2, li2024patchrefiner} have demonstrated significant potential for generating dense depth predictions from single images but still lack behind when delivering metric depth accuracy.
\PAR{Pseudo Segmentation for LiDAR data. }Recent advances in LiDAR pseudo-labeling leverage motion and appearance cues to generate robust labels, such as unsupervised instance segmentation methods to exploits these cues \cite{sautier2025unit} and methods extending 2D vision proposals into 3D space using grouping and voxelization techniques \cite{ošep2024bettersallearningsegment, gebraad2025leapconsistentmultidomain3d}. Additionally flow estimation through motion segmentation \cite{lin2024icp} can achieve real-time accuracy, but faces challenges with pose estimation in longer sequences.
\PAR{3D Pseudo Bounding Boxes for LiDAR Data. }Pseudo labeling has emerged as a pivotal technique in LiDAR object detection, addressing the reliance on extensive labeled datasets by generating pseudo labels for point clouds. 3DIoUMatch \cite{wang20213dioumatchleveragingiouprediction} employed a semi-supervised framework to filter high-quality pseudo labels object detection. More recently, \cite{Semoli} leveraged motion cues to group coherently moving points into objects, though tracking across numerous frames remains computationally demanding. Similarly, \cite{Baur2024ECCV} exploited self-supervised flow estimation and trajectory consistency to mine 3D bounding boxes. 
\PAR{Proposed Unified Labeling. }While addressing the same tasks tackled in isolation in prior work, we introduce a unified 3D labeling framework to concurrently deliver consistent depth estimation, object detection, and semantic segmentation pseudo labels, at longer range and without any form of supervision, as detailed in Table \ref{tab:checkmarks}. Despite handling three tasks together, our method still matches and surpasses dedicated models on each task.
\begin{figure*}[ht!]
\vspace{0mm}
    \centering
    \begin{center}
        \includegraphics[width=\textwidth]{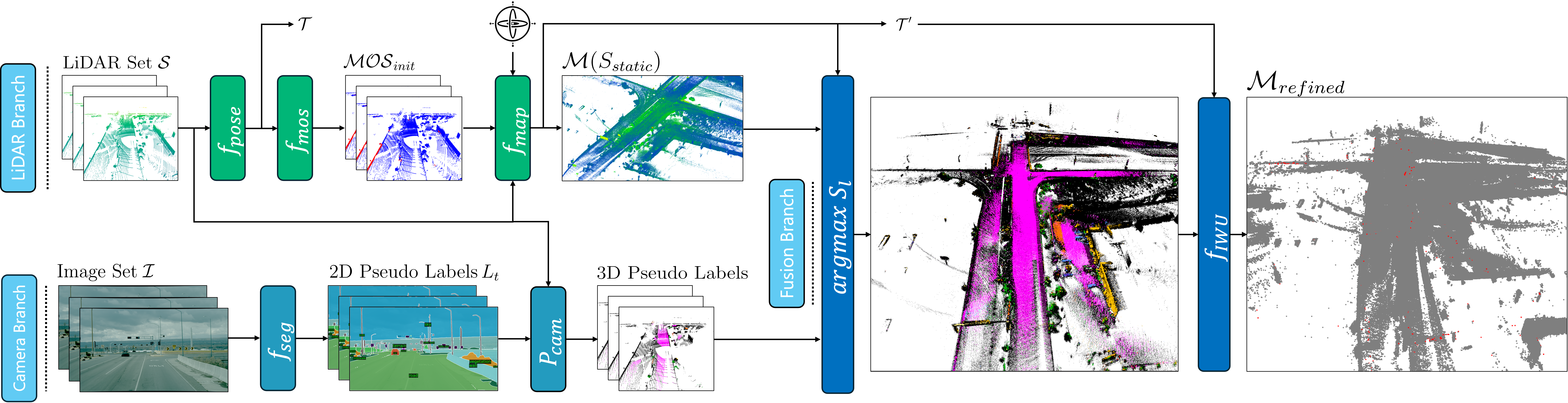}
    \end{center}
    \label{fig:mapLabeling}
        \vspace{-5mm}
    \caption{\textbf{Overview of Geometry-Grounded Dynamic Scene Decomposition.} Starting from a set of raw images, a set of LiDAR scan and IMU measurements, we first produce 3D semantic labels.
    Therefore, the 2D semantic masks produced by $f_{\text{seg}}$ are integrated into a map generated by the SLAM method $f_{\text{map}}$, by projecting them through $P_{\text{cam}}$, while simultaneously removing moving points identified by $f_{\text{mos}}$ from the map. To obtain a refined static scene map $\mathcal{M}_{refined}$ we first propagate the labels through geometric and temporal constraints and later on exploit them to remove remaining floaters and outliers (in red) through a our \textit{Iterative Weighted Update Function} $f_{IWU}$.}
    \label{fig:mapLabeling}
\end{figure*}
\begin{figure*}[ht!]
    \centering
    \begin{center}       \includegraphics[width=\textwidth]{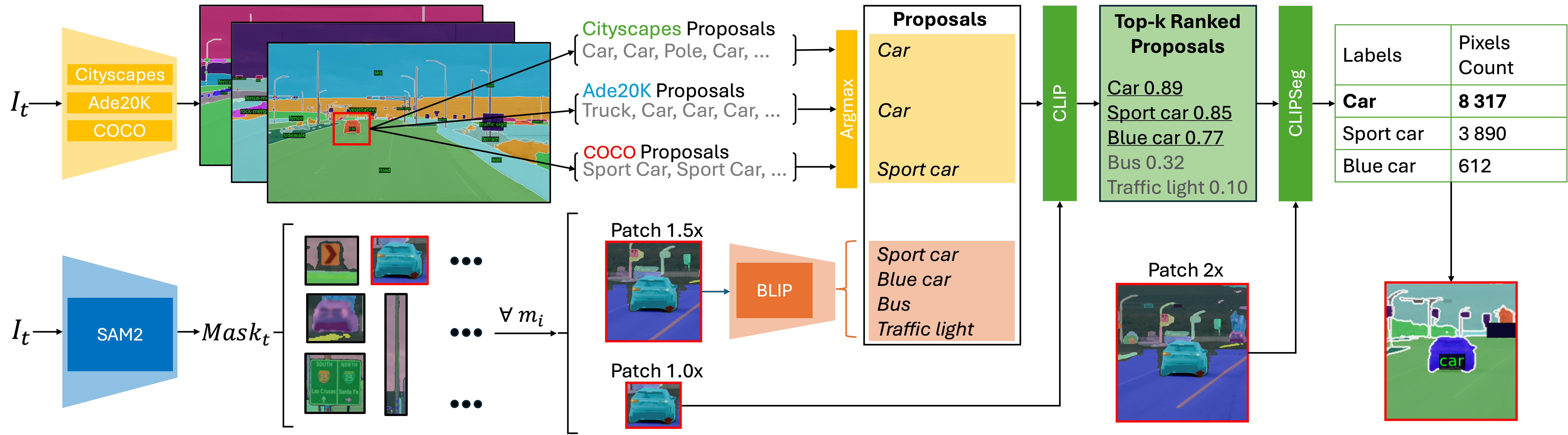}
    \end{center}
    \vspace{-5mm}
    \caption{\textbf{Overview of Pseudo Labeling Function.} Our proposed pseudo labeling method $f_{seg}$ robustly segments each $2D$ image $I_t$ by combining the predictions from an ensamble of three OneFormer \cite{jain2022oneformertransformerruleuniversal} models with weights from three different datasets (COCO~\cite{lin2015microsoftcococommonobjects}, ADE20K~\cite{zhou2017scene, zhou2019semantic} and Cityscapes~\cite{Cordts_2016_CVPR}) and a SAM2 \cite{ravi2024sam2} instance prediction set $Mask_t$. For each mask $m_i$, BLIP \cite{li2022blip} enriches the class proposals and through modal alignment with CLIP \cite{radford2021learningtransferablevisualmodels} and CLIPSeg \cite{lueddecke22_cvpr}, we ensure high quality domain-specific annotations.}
    \label{fig:pseudo-engine}
    \vspace{-5mm}
\end{figure*}

\section{Geometry-Grounded Pseudo-Labeling}
We introduce a pseudo-labeling method for LiDAR point clouds, agnostic to datasets and sensor setups, combining foundational vision models with geometry-aware probabilistic constraints on an accumulated scene map. As illustrated in Figure \ref{fig:mapLabeling}, images, LiDAR point clouds and IMU data are fused to initialize low-confidence semantic and motion labels. Iterating over the map we update labels probabilities, refining static structures and removing outliers through our \textit{Iterative Weighted Update}, yielding a dense $3D$ point map with reliable semantic labels and a sparse set of high-confidence moving objects.
\subsection{LiDAR Branch Processing }
We process a set of LiDAR scans $ \mathcal{S} = \{ s_t \mid t = 1, \dots, N \}$ and $IMU$ to obtain an accumulated map $\mathcal{M}(S_{static})$ free of the identified, low-confidence moving objects in $\text{MOS}_{\text{init}}$.
\PAR{Pose Estimation and Motion Cues.} We first apply a LiDAR‑odometry method $f_{\text{pose}}$ to estimate frame transformations $\mathcal{T}={T_t\in\text{SE}(3)}$, that a motion‑segmentation network $f_{\text{mos}}$ exploits to identify an initial set of low‑confidence moving objects $\text{MOS}_{\text{init}}$.
\par
\PAR{SLAM Mapping. }All points marked static are fused by a LiDAR‑inertial SLAM module $f_{\text{map}}$ to obtain the initial map $(\mathcal{M}, \mathcal{T}') = f_{\text{map}}(\mathcal{S}_{\text{static}}, IMU)$. Here
$\mathcal{S}_{\text{static}} = \{ s_t^{\text{static}} \mid s_t^{\text{static}} = s_t \setminus \text{mos}_t, \; s_t \in \mathcal{S}, \; \text{mos}_t \in \text{MOS}_{\text{init}} \}$. This early pruning of moving objects mitigates ghost artifacts and allows for better geometric optimization in the SLAM.
\subsection{Image Branch Processing}\label{sec:engine}
We process a set of images $\mathcal{I} = \{ I_t \mid t = 1, \dots, M \}$, to generate and lift $2D$ semantic pseudo labels in $3D$.
\PAR{Our pseudo labeling function} is presented in details in Figure \ref{fig:pseudo-engine}: given the set $\mathcal{I}$ of RGB images, for each $I_t\in\mathbb{R}^{H\times W\times3}$ it predicts a per‑pixel label image $L_t=f_{\text{seg}}(I_t)\in\mathcal{L}^{H\times W}$.
Each frame is down‑sampled and processed with SAM2 \cite{ravi2024sam2}, producing a set of object masks $Mask_t$.
The input image is segmented by three separate OneFormer \cite{jain2022oneformertransformerruleuniversal} models, individually trained on COCO~\cite{lin2015microsoftcococommonobjects}, ADE20K~\cite{zhou2017scene, zhou2019semantic}, and Cityscapes~\cite{Cordts_2016_CVPR}. For each $m_i\!\in\!Mask_t$, the initial proposal list is generated by stacking the most recurrent ($Argmax$) class from each OneFormer model per-pixel logit maps. We then extract three image crops centered on the mask’s bounding box: original size ($1.0\times$), large ($1.5\times$), and huge ($2.0\times$). Subsequently, open-vocabulary classification is performed by non-prompted BLIP \cite{li2022blip} on the large patch, proposing class candidates that augment the OneFormer proposal list. CLIP~\cite{radford2021learningtransferablevisualmodels} re‑ranks the candidates list on a tighter crop, producing a shortlist of the top-k keywords: we select specifically $k=3$. CLIPSeg~\cite{lueddecke22_cvpr} processes the full‑resolution crop together with this shortlist and outputs per‑pixel scores. A majority vote assigns the final class to all pixels of $m_i$, reducing boundary noise. If multiple classes remain, we keep the one with the highest pixel count. Iterating through every $m_i\!\in\!Mask_t$ we obtain a refined label map $L_t$ which serves as initial guess.
\PAR{Occlusion Aware Semantic Lifting. }\label{par:vismask}Each LiDAR pointcloud $s_t$ is projected into the correspondent label map $L_t(u,v)$ with calibration matrix $P_{cam}$. Let $D(u,v)$ be the depth at pixel $(u,v)$, the visibility mask $ M(u, v) $ is determined by comparing the depth with the minimum depth in its neighborhood $ \mathcal{N}(u, v) $. A point is marked as visible if
\begin{equation}
    D(u, v) \leq \min(D(\mathcal{N}(u, v))) + 0.5 .
\end{equation}
Only visible points inherit the semantic label $l_t(u,v)$ of $L_t(u,v)$, ensuring noisy labels are reduced in sparser region or in common penetration cases.
\subsection{Geometry-Consistent Fusion Branch}
After differentiating the world into static world ($\mathcal{M}$) and dynamic objects ($\text{MOS}_{\text{init}}$) we propose a geometry-grounded method to iteratively refine the static world representation.
\PAR{Semantic Multimodal Propagation.}\label{par:semmapprop} By sequentially associating each point label of a LiDAR scan to the correspondent point in the map, we project the semantics from each camera into the world map. As a result, each point in the map then is be represented as $p_i = \left( x_i, y_i, z_i, \left\{ \left( l_{i1}, n_{i1} \right), \left( l_{i2}, n_{i2} \right), \dots, \left( l_{im_i}, n_{im_i} \right) \right\} \right)$, where $x_i, y_i, z_i$ are the spatial coordinates and $\left\{ \left( l_{ij}, n_{ij} \right) \right\}$ is a set of label-count pairs associated with point $i$, and
\begin{itemize}
    \item $l_{ij}$ is the $j$-th label assigned to point ${p}_i$.
    \item $n_{ij}$ is the number of times label $l_{ij}$ was assigned ${p}_i$.
\end{itemize}
Here, $m_i$ is the total number of unique labels assigned to point $i$. We propagate labels probabilistically in order to enhance segmented areas and fill gaps in the map following Algorithm 
\ref{alg:labelrefinement}.
\begin{algorithm}[t]
\caption{Probabilistic Label Propagation}\label{alg:labelrefinement}
\begin{algorithmic}[1]
\Require Point cloud $\{ (p_i, l_i) \}_{i=1}^N$, neighborhood radius $r$
\Ensure Refined labels $\{ l_i \}_{i=1}^N$
\State Build a KD-Tree from points $\{ p_i \}$
\State Set $\sigma = \dfrac{r}{2}$
\ForAll{points $p_i$}
    \State $\mathcal{N}_i = \{ (p_j, l_j) \mid \| p_i - p_j \| \leq r, \ l_j \notin \{0, -1\} \}$
    \If{$\mathcal{N}_i \neq \emptyset$}
        \State 
        $S_l = \sum_{(p_j, l_j) \in \mathcal{N}_i} w_{ij} \cdot \delta(l_j = l)$
        \State
        $l_i = \underset{l}{\operatorname{arg\,max}} \; S_l$
        
    \EndIf
\EndFor
\end{algorithmic}
\end{algorithm}
Here, $w_{ij} = \exp\bigl(-\|p_i - p_j\|^{2}/(2\sigma^{2})\bigr)$, and $\delta(l_j = l)$ equal to 1 if $l_j = l$, and 0 otherwise.
\PAR{Map Refinement.}
To refine the map from remaining floaters we propose our \textit{Iterative Weighted Update Function} $f_{IWU}$: by iteratively comparing the sparse LiDAR with the map, points belonging to moving objects but mistakenly registered in the map are likely to be observed only once or twice by subsequent scans. Consequently, we update the probability that each map point is static by considering the frequency of its observations, incorporating a distance based influence factor. 
For each point $ {p}_j \in {s}_t, {s}_t \in \mathcal{S}, t = 1, \dots, N $, we calculate the Euclidean distance $ {d}_{ij} $ to all map points $ {m}_i \in \mathcal{M}$ and from the sensor origin $ {r}_j $, $d{ij} = \|{p}_j - {m}_i\| , \quad  r_j = \|{p}_j\|$. We locate the nearest map point $ {\tilde{m}}_i $ for each scan point $ {p}_j $ and compute
\begin{equation}
    {r^*}_j = \min\left(1, \frac{r_{\text{max}}}{{r_j}}\right),\quad
    C(\tilde m_i)=\frac{\max_j n_{ij}}{\sum_j n_{ij}} ,
\end{equation}
with \(n_{ij}\) counting how often \(\tilde m_i\) was labeled as class \(l\in\{\text{movable},\text{non-movable}\}\) and $r_{max}$ defining a full-credibility radius of $200$ meters.
The static probability update rule, if $ {\tilde{m}}_i $ is found in a $30$ centimeters radius, is:
\begin{equation}
\small
P^{t}(\tilde m_i)=\alpha \cdot P^{t-1}(\tilde m_i)+(1-\alpha)\cdot r_j^{*}\cdot\bigl(1+C(\tilde m_i)\bigr) ,
\label{eq:update-in}
\end{equation}
otherwise
\begin{equation}
\small
P^{t}(\tilde m_i)=\alpha \cdot P^{t-1}(\tilde m_i)+(1-\alpha)\cdot (1 - r^*_j)\cdot\bigl(1-C(\tilde m_i)\bigr).
\label{eq:update-in}
\end{equation}
Points with probabilities exceeding a predefined threshold $\tau_s$ are classified as static, while those below $\tau_s$ are marked as moving and discarded from the map.
\subsection{Pseudo Label Outputs}\label{sec:pseudooutputs}
\begin{figure*}[ht!]
\vspace{0mm}
    \centering
    \begin{center}
        \includegraphics[
            trim=0.4cm 0cm 0cm 0cm,  
            clip,
        width=\textwidth]{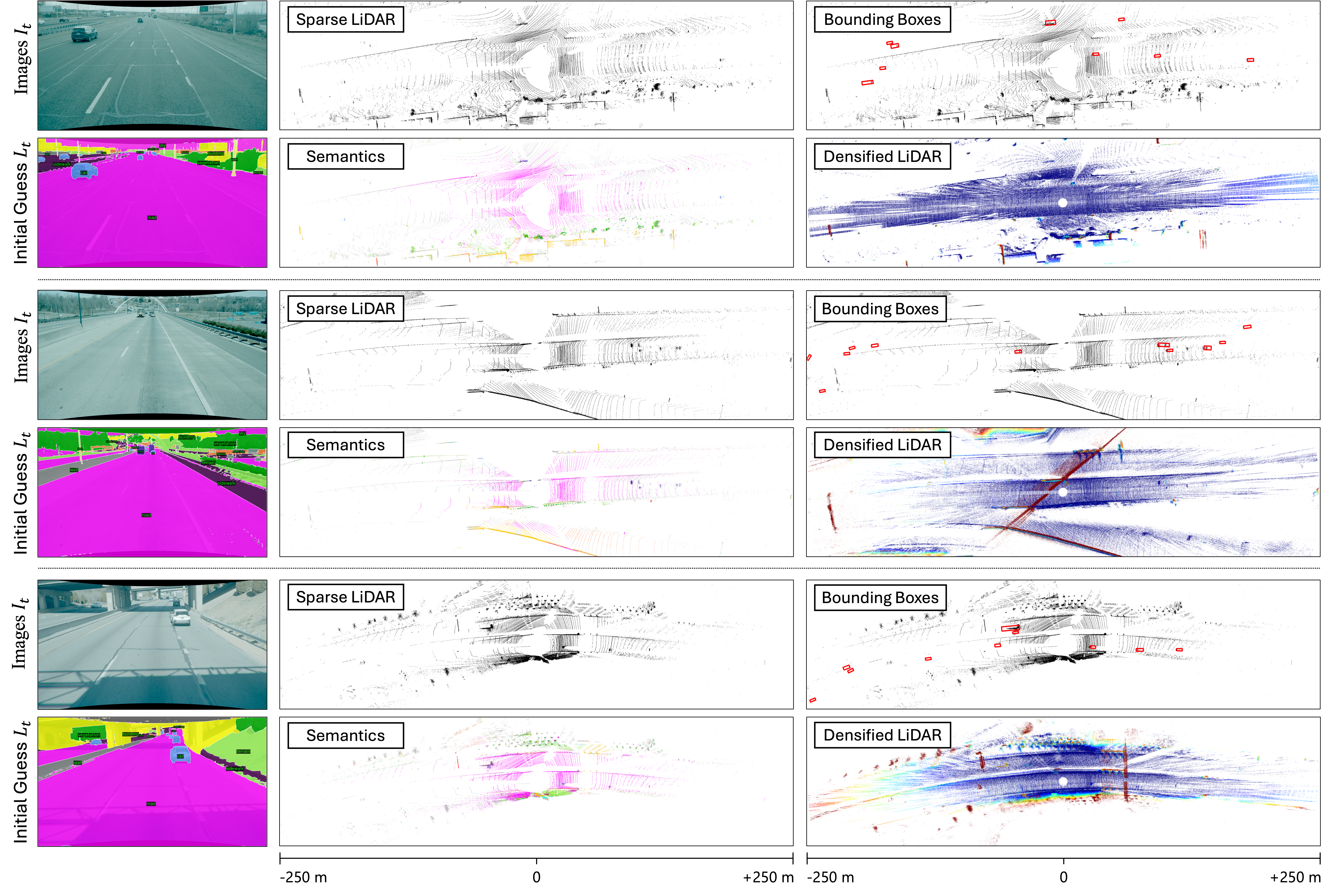}
    \end{center}
    \vspace{-7mm}
    \caption{\textbf{UniLiPs Unified Labeling Outputs. }Coupling geometric point-cloud aggregation with image segmentation cues from our $f_{seg}$, UniLiPs rivals standalone methods by jointly producing temporally consistent semantic labels, trajectory-smoothed bounding boxes, and densified LiDAR sweeps that are denser and offer finer angular resolution, especially at long range. In the Figure, densified LiDARs are z-colored between $-2$ (blue) and $+5$ (red) meters, while semantics are class-coloured based on SemanticKITTI mapping.}
    \label{fig:outputs}
    \vspace{-5mm}
\end{figure*}

After the refinement stage, our pipeline can generate different pseudo ground‑truth supervision signals like densified LiDAR scans, 360° semantic labels and $3D$ bounding boxes from moving‑object segmentation masks, as in Figure~\ref{fig:outputs}.
\PAR{3D Semantics.} We extract semantic labels for each LiDAR scan from the semantically propagated map, preserving the initial guess $l_{ij}$ for points without correspondence.
\PAR{Moving Objects.} We detect moving objects by aligning each LiDAR scans in $\mathcal{S}$ to the refined consistent static map $\mathcal{M_\text{refined}}$, and segment as moving those points without correspondence in the map, requiring the existence of at least $2$ other moving candidates in a neighborhood of $1$ meter. 
\PAR{3D Bounding Boxes}. To transform the moving object detections into bounding boxes, we first exploit the pose $\mathcal{T}'$ to align three consecutive scans, considering only the points labeled as moving, and then use HDBSCAN \cite{mcinnes2017hdbscan} to cluster them. We fit the minimum enclosing cuboid to each cluster and assign a minimum size; we use PCA to get an initial estimate of the yaw and use a Kalman Filter based tracker, with constant velocity model including yaw dynamics. We then refine each object trajectory using a spline optimization method. We represent the yaw \( \psi \) as a combination of basis functions, modeling both the sine \( f_{s}(t) \) and cosine \( f_{c}(t) \) components, and minimize the following cost function,
\begin{equation}
  e_{\text{yaw}}=\tfrac12\sum_{j}
     \bigl((f_c(t_j)-\cos\psi_j)^2+(f_s(t_j)-\sin\psi_j)^2\bigr)
\end{equation}
We then employ the $x$ and $y$ positions and minimize
\begin{equation}
e_{\text{position}} = \frac{1}{2} \sum_{j} \left( \left( f_x(t_j) - x_j \right)^2 + \left( f_y(t_j) - y_j \right)^2 \right)
\end{equation}
where $x_j$ and $y_j$ represent the measured positions at time $t_j$.
More details in the Supplementary Material.
\par

\PAR{High-quality Accumulated Depth}. Further we provide high resolution LiDAR frames  from the accumulated map, exploiting the pose $\mathcal{T}'$ to reintroduce moving objects corresponding to that specific pose and time, and to transform the coordinate system. To compensate for occlusions our \textit{Adaptive Spherical Occlusion Culling} converts each point to spherical coordinates $(r, \theta, \phi)$, define angular resolutions $\Delta\theta$ and $\Delta\phi$, and create bins
\begin{linenomath*}
  \begin{equation}
    \theta_{\text{bins}} = \left\{ \theta_{\min}, \theta_{\min} + \Delta\theta, \theta_{\min} + 2\Delta\theta, \dots, \theta_{\max} \right\}
  \end{equation}
\end{linenomath*}
\begin{linenomath*}
  \begin{equation}
    \phi_{\text{bins}} = \left\{ \phi_{\min}, \phi_{\min} + \Delta\phi, \phi_{\min} + 2\Delta\phi, \dots, \phi_{\max} \right\}
  \end{equation}
\end{linenomath*}
In contrast to existing methods, for each bin $(i, j)$, we find the minimum range $r_{\min}^{(i, j)}$ , that is
\begin{equation}
r_{\min}^{(i, j)} = \min \left\{ r_k \mid k \in \text{bin } (i, j) \right\} .
\end{equation}
and define a threshold function $T(r)$ that increases with range $T(r) = 1 + \alpha r$, where $\alpha$ is a small positive constant. A point $k$ in bin $(i, j)$ is considered visible if
$r_k \leq r_{\min}^{(i, j)} + T(r_k)$ and otherwise, the point is considered occluded. These high-resolution LiDAR frames, with three to five times the density across all ranges, serve as reference data (ground truth) for depth learning from images.
\par
\section{Experiments}
To validate our approach, which is unique in its ability to generate pseudo-labels simultaneously for $3$ tasks, we benchmark it against state-of-the-art methods that tackle each task in isolation. We conduct experiments on the short-range datasets KITTI \cite{geiger2012we} and nuScenes \cite{caesar2020nuscenes}, and an experimental long-range highway dataset that captures beyond the $80m$ LiDAR range limit of public datasets. Adhering to the common protocol adopted in recent works \cite{Semoli, lin2024icp, kong2023lasermix, gebraad2025leapconsistentmultidomain3d}, we first compare our pseudo-labels with human annotations and then train task-specific models on a mix of ground-truth and pseudo-labels. This evaluation highlights both the accuracy of our pseudo annotations and the extent to which models can absorb the noise introduced by pseudo-labeling.
\subsection{Common Settings}
Across all datasets used, we kept the parameters necessary for our method constant: we set the label propagation radius $r=0.2m$, the probability threshold for moving points detection to $\tau_s=0.5$ and for the probability update $\alpha=0.7$. 
\subsection{Accumulated Pseudo-Depth Evaluation}
We evaluate pseudo-depth generation by finetuning an NMRF \cite{guan2024neural} model, using both the short-range-LiDAR, small-baseline stereo pairs on KITTI \cite{geiger2012we} and our long-range-LiDAR, and wide-baseline cameras. We supervise NMRF with projected pseudo LiDAR and reverse Huber loss \cite{laina2016deeper, zwald2012berhupenaltygroupedeffect}, and validate the improvements computing Mean Absolute Error (MAE) and Root Mean Squared Error (RMSE) on the pixels where ground truth is available.
\par
\PAR{Baselines. }To compare our pseudo depth generation with existing methods, we produce pseudo ground-truth depth using three distinct baselines. First, a LiDAR-based, dense pseudo depth obtained through LIO-SAM \cite{liosam2020shan}, to show the importance of our Adaptive Spherical Occlusion Culling and floaters refinement. Second, a monocular foundation model \cite{depth_anything_v2} to predict metric depth from single images. Third, a robust stereo prediction network \cite{li2022practical} to generate depth maps. The resulting pseudo-labeled frames are used for depth supervision following a consistent train-test split.
\par
\begin{table}[!t]
    \centering
    \small
    \setlength{\tabcolsep}{2.5pt}
    \begin{tabular}{lc| ccc| ccc}
        \toprule
         & \textbf{Pseudo} 
        & \multicolumn{3}{c}{\textbf{MAE $\downarrow$ [m]}}
        & \multicolumn{3}{c}{\textbf{RMSE $\downarrow$ [m]}} \\
        \cmidrule(lr){3-5}
        \cmidrule(lr){6-8}
        &
        & 0-80 & 80-150 & 150-250
        & 0-80 & 80-150 & 150-250 \\
        \midrule
        \multirow{5}{*}{\rotatebox{90}{\textbf{KITTI}}}
& \cellcolor{verylightgray} Oracle
 & \cellcolor{verylightgray} {4.48} & \cellcolor{verylightgray} 22.03 & \cellcolor{verylightgray} 30.76
 & \cellcolor{verylightgray} {7.62} & \cellcolor{verylightgray} 25.66 & \cellcolor{verylightgray} 35.83
 \\
& LIO-SAM
 & \underline{4.71} & {\underline{13.00}} & {\underline{18.16}}
 & {8.25} & {\underline{16.67}} & {\underline{22.55}}
  \\
& CREStereo
 & 8.19 & 17.72 & 23.90
 & 10.99 & 21.62 & 27.05
 \\
& DA-V2
 & 6.38 & 15.73 & 22.17
 & \underline{8.00} & 21.61 & 29.98
\\
& \textbf{Proposed}
 & {\textbf{3.28}} & {\textbf{9.57}} & {\textbf{17.43}}
 & {\textbf{5.66}} & {\textbf{13.49}} & {\textbf{21.89}}
 \\
\cline{1-8}\noalign{\vskip 0.8mm}
\multirow{5}{*}{\rotatebox{90}{\textbf{Long Range}}}
& \cellcolor{verylightgray} Oracle
 & \cellcolor{verylightgray} {5.44} & \cellcolor{verylightgray} 20.79 & \cellcolor{verylightgray} {31.83}
 & \cellcolor{verylightgray} {7.98} & \cellcolor{verylightgray} 25.70 & \cellcolor{verylightgray} {38.96}
 \\
& LIO-SAM
 & \underline{5.53} & {\underline{11.89}} & \underline{32.33}
 & \underline{10.51} & {\underline{20.11}} & 46.22
 \\
 & CREStereo
 & 8.33  & 22.34 & 37.04  
 & 10.90 & 27.13 & 45.86  
 \\
 & DA-V2
 & 8.31 & 23.55 & 32.67  
 & 11.83 & 28.75 & \underline{39.96}
 \\
 & \textbf{Proposed}
 & {\textbf{2.27}} & {\textbf{6.14}} & {\textbf{21.07}}
 & {\textbf{4.21}} & {\textbf{9.81}} & {\textbf{25.16}}
 \\
        \bottomrule
    \end{tabular}
    \vspace{-2mm}
    \caption{\textbf{Depth Estimation Evaluation} of NMRF \cite{guan2024neural}, supervised with pseudo depth frames from LiDAR and Image methods, on KITTI and our Long Range Dataset. Excluding the Oracle (in gray), best results are \textbf{bold}; second bests are \underline{underlined}.}
    \label{tab:results_depth}
    \vspace{-5mm}
\end{table}
\PAR{Short Range Dataset (KITTI). }We randomly sample the sequences from the KITTI training dataset to obtain training and evaluation sets. 
Then, we evaluate the NMRF model, by using the weights pretrained on synthetic data~\cite{MIFDB16} and fine-tuned on a subset of the naively sparse LiDAR (denoted as \textit{Oracle}).
For all other methods we sub select a set of $400$ stereo pairs and train for $30,000$ steps on our accumulated depth as well as on the introduced alternative sources of gt-depth data. The performance is evaluated against the pseudo ground truth generated from our accumulated LiDAR data as this allows us to evaluate ranges up to 250m. Evaluations of standard KITTI ranges are shown in the Supplementary Material. 
Final results are reported in Table \ref{tab:results_depth}, where we observe an improvement of $26.8\%$, $56.6\%$ and $43.3\%$ in MAE and $25.7\%$, $47.4\%$, $38.9\%$ in RMSE for $0$-$80$m, $80$-$150$m, $150$-$250$m ranges respectively. Moreover, we achieve an average improvement over all baselines of  $46.3\%$, $37.2\%$, and $17.5\% $in MAE and $36.4\%$, $31.4\%$, and $16.3\%$ in RMSE for $0$-$80$m, $80$-$150$m, $150$-$250$m ranges, respectively.
\PAR{Long-Range Dataset.}
We generate $400$ ground truth samples for training extracted from diverse highway scenes. For a fair comparison, we first fine-tune the pre-trained model on our sparse LiDAR recordings, as our sensor can capture points at longer ranges compared to the Velodyne HDL-64E deployed in KITTI \cite{geiger2012we}, with the same number of iterations used later for the dense ground truths. Then, we fine-tune on each of the aforementioned pseudo-ground truths.
The results are reported in Table \ref{tab:results_depth}, where we observe an improvement of $58.7\%$, $70.2\%$ and $33.2\%$ in MAE and $47.6\%$, $61.3\%$, $35.7\%$ in RMSE for $0$-$80$m, $80$-$150$m, $150$-$250$m ranges respectively.
Moreover, we achieve an average improvement against all baselines of $68.1\%$, $64.9\%$ and $37.8\%$ in MAE and $61.9\%$, $60.3\%$, and $42.6\%$ in RMSE for $0$-$80$m, $80$-$150$m, $150$-$250$m ranges respectively.
Especially on our dataset, rich in highway scenarios, reference SLAM system often encounter numerous dynamic objects that leave residual traces, or "floaters" (showed qualitatively in the Supplement), which degrade the accuracy of depth predictions and \textit{confirm that our refinement method significantly enhances performance by effectively reducing these inaccuracies}. 
\subsection{Semantic Pseudo-Labels Evaluation}
We evaluate our semantic pseudo-labels on SemanticKITTI \cite{behley2019iccv} \textit{val} sequence $08$ and on more than $40\text{k}$ samples of NuScenes~\cite{caesar2020nuscenes}, generating them using only the front-left camera for the former and all six cameras for the latter.
\par
\begin{table}[!ht]
    \centering
    \setlength{\tabcolsep}{+3pt} 
    \small
    \begin{tabular}{lccc|cc}
        \toprule
        
         & & \multicolumn{2}{c}{\textbf{KITTI}} 
        & \multicolumn{2}{c}{\textbf{nuScenes}} \\
        \cmidrule(lr){3-4}\cmidrule(lr){5-6}
        & \textbf{Method} & \textbf{mIoU} & \textbf{cat.mIoU} 
                       & \textbf{mIoU} & \textbf{cat.mIoU} \\
        \midrule
        \multirow{4}{*}{\rotatebox{90}{\textbf{POINT}}} &
        SSAM
            & 10.7 & 19.7 & \underline{13.4} & \underline{23.2} \\
        & LeAP (points) 
            & \underline{46.8} & \underline{68.6} & -- & -- \\
        & \textbf{Our} $\mathbf{f_{\text{seg}}}$
            & \textbf{59.4} & \textbf{69.6} & \textbf{54.9} & \textbf{62.6} \\ 
        \cline{2-6}\noalign{\vskip 0.8mm}
        & \textbf{Our (Propagated)}
            & {64.9} & {76.2} & {58.0} & {65.2} \\ 
        \midrule
        \multirow{3}{*}{\rotatebox{90}{\textbf{VOXEL}}} &
        SSAM (Propagated)
            & 11.5 & 23.9 & \underline{25.3} & \underline{29.2} \\
        & LeAP + 3D-CN (2) 
            & \underline{58.1} & \underline{81.6} & -- & -- \\
        & \textbf{Our (Propagated)}                    
            & \textbf{68.3} & \textbf{86.6} & \textbf{59.1} & \textbf{76.3} \\ 
        \bottomrule
    \end{tabular}
    \vspace{-2mm}
     \caption{\textbf{Pseudo Labels SOTA Comparison.} We evaluate pseudo labels generated by our $f_{\text{seg}}$ and the refined ones on Semantic KITTI and NuScenes, on the \cite{9786676} benchmark reduced sets of classes, per-point and voxelizing, according to LeAP \cite{gebraad2025leapconsistentmultidomain3d}. Best results are \textbf{bold}; second bests are \underline{underlined}.}
    \label{tab:leap}
    \vspace{-4mm}
\end{table}
\PAR{Semantic Pseudo Labels Comparison.} 
We evaluate and compare our pseudo labels against LeAP \cite{gebraad2025leapconsistentmultidomain3d} and Semantic SAM \cite{chen2023semantic}: for a fair comparison we as well consider only labeled points and reduce the number of classes to a set of $11$ classes (\textit{car, bicycle, motorcycle, other-vehicle, person,road, sidewalk, other-ground, manmade, vegetation, terrain}) as well as to the 6 coarse \textit{category} classes (\textit{flat, construction, object, nature, human, vehicle}), both well defined in the benchmark paper of KITTI 360 \cite{9786676}. Results shown in Table \ref{tab:leap} highlight how our pseudo labeling function \( f_{\text{seg}} \) (§ \ref{sec:engine}) is the most accurate in labeling LiDAR data. To further compare with LeAP, which propagates its initial labels on a $0.2 m$ voxel grid, we voxelize our propagated labels at the same $0.2 m$ resolution and compare with the reported best result. Thanks to the higher point-level accuracy of our propagation technique (in Table~\ref{tab:leap} point propagated), our voxelized predictions outperform all competing methods without the need for additional voxel refinement, achieving state of the art in semantic pseudo-labeling.
\par
\PAR{Quality vs. Oracle. }
We select PVKD~\cite{pvkd} as fixed off-the shelf model to be trained: we keep all hyper-parameters fixed, and vary only the supervision source. In the \emph{Oracle} case we use $100$\,\% Semantic KITTI ground-truth labels; for \emph{Limited GT} a randomly chosen $10$\% ground-truth subset; for \emph{Our} we feed that identical $10$\,\% subset plus $90$\,\% pseudo labels generated by our pipeline.  Each regime is repeated five times with different $10$\,\% splits, and mIoU on \textit{val} sequence 08 is reported in Table~\ref{tab:comparison_pseudo_labels}.  Our pseudo labels recover near-oracle performance, with a small average difference of $1.09\%$ mIoU  and of $0.30\%$ when classes not predicted by our method (parking, bicyclist, motorcyclist, other-ground, other-objects, trunk) are excluded. To compare these results, we train the same PVKD network with pseudo labels supervision from three alternative sources: 3D projections of Semantic SAM predictions with temporal propagation (\emph{SSAM})~\cite{chen2023semantic}, the self-supervised LaserMix training scheme~\cite{kong2023lasermix} and inference pseudo-labels from a Cylinder3D~\cite{zhu2021cylindricalasymmetrical3dconvolution} model pre-trained on nuScenes~\cite{caesar2020nuscenes} and  lightly fine-tuned for two epochs on $2 000$ SemanticKITTI frames with $1/10$ the learning rate. Across all the comparisons, the PVKD model trained on our labels delivers consistently higher mIoU than when trained on baselines pseudo labels, requiring \emph{no} extra manual annotation, underscoring their effectiveness for semantic oracle recovery. Moreover, aside from Semantic SAM, which achieves rather weak performance, our approach is the only one that can function  ($0$-$100$), entirely without ground-truth supervision. The strongest competing baseline, LaserMix, can work with small proportions of ground-truth in the GT-pseudo mix, yet it still needs some labeled data and \textit{cannot handle the $0 \%$ ground-truth regime} that our method successfully addresses.
\begin{table}[t]
\centering
\setlength{\tabcolsep}{+2pt}
\renewcommand{\arraystretch}{1.2}
\resizebox{\linewidth}{!}{
\begin{tabular}{lccccc} 
\hline
\multirow{2}{*}{\textbf{Supervision}} & \multirow{2}{*}{\shortstack{\textbf{GT} \\ \textbf{Pseudo}}} & \multicolumn{2}{c}{\textbf{All Classes}} & \multicolumn{2}{c}{\textbf{Mapped Classes}} \\
\cline{3-6}
 & & mIoU\% &  Oracle \% & mIoU\% & Oracle \%\\
\hline
Limited GT & 10-0                      & 43.41 & 70.3  & 48.25 & 70.3  \\
\rowcolor{verylightgray} Oracle & 100-0             & 61.73 & - & 68.63 & - \\
SSAM \cite{chen2023semantic} & 10-90 & 33.70 & 54.6  & 43.17 & 63.0  \\
Pre-Trained \cite{zhu2021cylindricalasymmetrical3dconvolution} & 10-90 & 44.87 & 72.7 & 52.07 & 75.9 \\ 
Lasermix Vx \cite{kong2023lasermix} & 10-90 & {59.38} & {96.3}  & \underline{67.49} & \underline{98.4}  \\
\cline{1-6}\noalign{\vskip 0.8mm}
{UniLiPS} Full (ours) & 0-100                      & 51.48 & 83.4  & 55.10 & 80.3  \\
{UniLiPS} $95\%$ (ours) & 5-95                      & \underline{59.46} & \underline{96.3}  & 66.71 & 97.2  \\
\textbf{UniLiPS} $90\%$ (ours) & 10-90 & \textbf{60.63} & \textbf{98.2}  & \textbf{68.33} & \textbf{99.6}  \\
\hline
\end{tabular}}
\vspace{-3mm}
\caption{\textbf{Semantic Segmentation.} We evaluate pseudo labels quality supervising a PVKD \cite{pvkd} model with pseudo labels produced by different methods. Our results demonstrate that incorporating additional pseudo-labels is crucial for regaining oracle-level performance, as evidenced by the differences between the $10-0$ and $10-90$ configurations. Furthermore, our approach benefits from label re-weighting and accumulation, yielding significant improvements over the Semantic SAM baseline. Excluding Oracle (in gray), best results are \textbf{bold}; second bests are \underline{underlined}.}
\label{tab:comparison_pseudo_labels}
\vspace{-3mm}
\end{table}
\subsection{Object Detection Evaluation}
We evaluate our pseudo bounding boxes performance on our highway long range dataset.
\PAR{Pseudo Bounding Boxes} are evaluated in Table \ref{tab:ODperformance} using mAP and ND-Score, with $6$ meters threshold, on a maximum range of $250$ meters. We compare with LISO \cite{Baur2024ECCV}, due to the similar detection-trajectory-refinement methodology. We train their model on our data and produce inference bounding box on the same validation split. Additionally, we compare against pseudo bounding boxes from ICP-Flow \cite{lin2024icp}, an effective annotation-free pseudo-labeling method: we threshold its flow estimates at $1$ m/s to segment movers, then derive boxes using our procedure.
\par
\begin{table}[t]
\centering
\begingroup
\small  
\setlength{\tabcolsep}{12pt} 
\renewcommand{\arraystretch}{1.0}   
\begin{tabular*}{0.95\linewidth}{@{\extracolsep{\fill}} l c c c @{}}
  \toprule
  & ICP\textendash{}Flow~\cite{lin2024icp} & LISO~\cite{Baur2024ECCV} & \textbf{Ours} \\
  \midrule
  \textbf{mAP [\%]} & 7.2 & \underline{21.1} & \textbf{31.0} \\
  \textbf{NDS [\%]} & 11.4 & \underline{40.9} & \textbf{45.2} \\
  \bottomrule
\end{tabular*}
\endgroup

\vspace{-3mm}
\caption{\textbf{Pseudo Bounding Boxes} evaluation on the highway-driving dataset. We achieve state-of-the-art compared to other pseudo-labeling and detection-from-motion approaches.}
\label{tab:ODperformance}
\vspace{-1mm}
\end{table}
\PAR{Quality Vs Oracle.} Secondly, we train an off-the-shelf $3D$ detector following the architecture of PointPillars\cite{lang2019pointpillars} on full ground truth (\textit{Oracle}) and on $20\%$ ground truth and $80\%$ pseudo labels, generated by our method (\textit{Proposed}) and by using ICP-Flow, as described before. We report the results in Table \ref{table:ODres}, where we find our pseudo labels can achieve near-oracle performances compared to other effective pseudo labeling methods.
\par
\begin{table}[t]
\centering
\setlength{\tabcolsep}{2.1pt}
\resizebox{\linewidth}{!}{
\begin{tabular}{l | c c | c c | c c}
\hline
\multirow{2}{*}{\textbf{Method}} & \multicolumn{2}{c|}{\textbf{-25 - 25m}} & \multicolumn{2}{c|}{\textbf{-50 - 50m}} & \multicolumn{2}{c}{\textbf{-70 - 70m}} \\
 & bev AP & 3d AP & bev AP & 3d AP & bev AP & 3d AP \\
\hline
    \rowcolor{verylightgray} \textbf{Oracle} & 35.55 & 34.45 & 33.54 & 33.08 & 32.44 & 30.13 \\
    \textbf{ICP-Flow \cite{lin2024icp}} & 11.01 & 3.60 & 9.45 & 3.41 & 9.44 & 3.20 \\
    \textbf{Proposed} & \textbf{31.02} & \textbf{26.53} & \textbf{29.43} & \textbf{25.44} & \textbf{29.19} & \textbf{25.33} \\
\hline
\end{tabular}}
\vspace{-3mm}
\caption{\textbf{Object Detection Evaluation Results} on the challenging experimental highway dataset: the model trained on our pseudo labels achieves near-\textit{Oracle} performances compared to baseline methods. Excluding Oracle (greyed out), best results are \textbf{bold}.}
\vspace{-1mm}
\label{table:ODres}
\end{table}
\section{Ablations} 
\begin{figure}[t]
    \centering
       \includegraphics[
            trim=0cm 0.4cm 0.3cm 0.3cm,  
            clip,
            width=\linewidth]{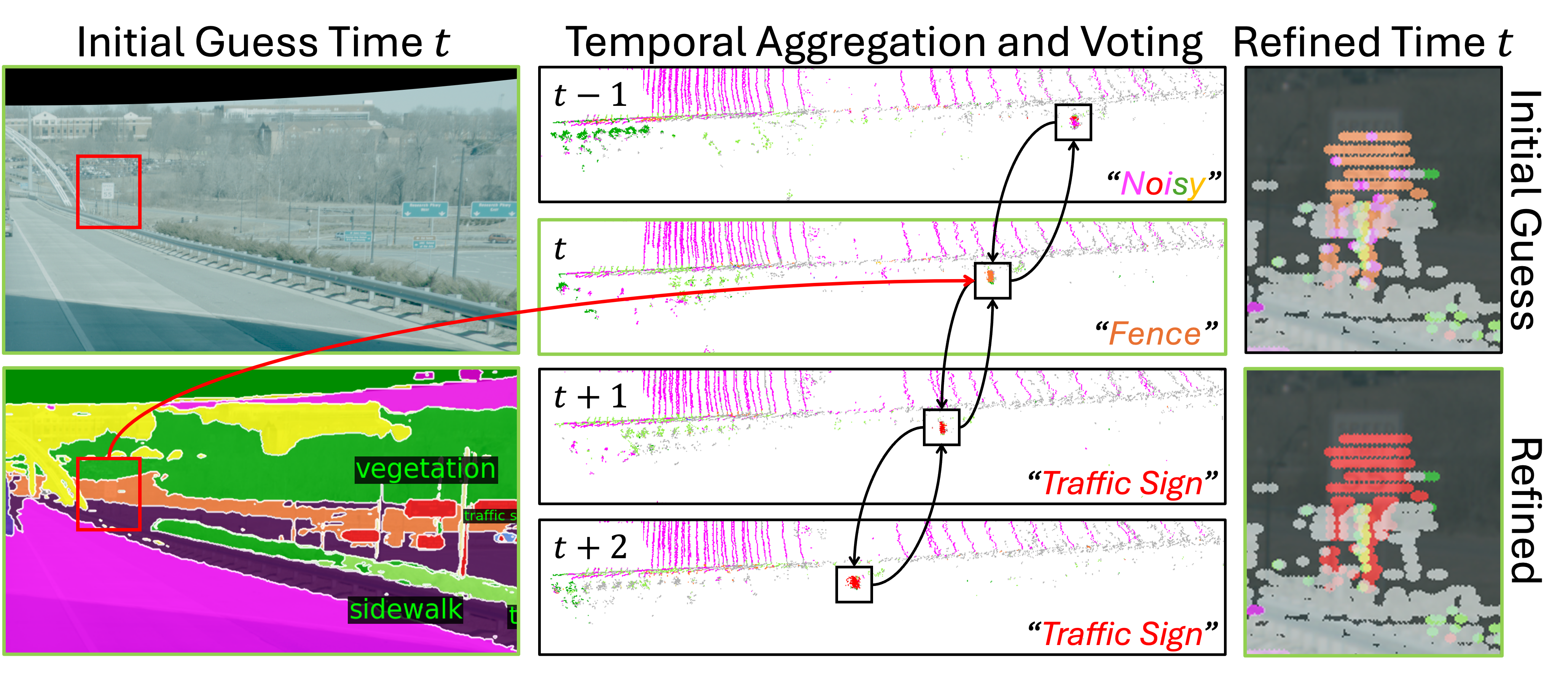}
    \vspace{-6mm}
    \caption{\textbf{Effect of Semantic Multimodal Propagation.} Leveraging our refined, geometry-grounded map as a reference, mislabeled points in each LiDAR scan are systematically corrected, ensuring label consistency across all timestamps.
    }
    \label{fig:ablation}
    \vspace{-2mm}
\end{figure}

Figure~\ref{fig:ablation} shows qualitatively the importance of our geometry grounded label propagation for temporal consistency and reweighting of mislabeled points.
Table \ref{tab:pseudo_labels_ablation} reports point-wise mIoU after ablating each $f_{seg}$ sub-model, highlighting their individual impact. Additionally, we note that increasing the number of top-k CLIP proposals ($k > 3$) doesn´t impact the mIoU score on the evaluated datasets. In Table~\ref{tab:spline_opt_abl} we analyze performance drop of our pseudo bounding boxes after ablating the spline optimization, which helps pose and orientation score, and our $f_{iwu}$, which increases detection probability. In Table \ref{tab:abl_occ_mask} we complement Table~\ref{tab:leap} point-wise evaluation ablating sequentially Algorithm \ref{alg:labelrefinement}, the accumulation and the occlusion mask in the lifting module (§\ref{par:vismask}): the former effectively re-weights labels, especially in dense regions, for more accurate prediction, while the latter removes noise from penetration and misaligned projections. More ablations are presented in the Supplement. 
\par
\begin{table}[t]
\addtolength{\tabcolsep}{-0.15em}
\small
\centering
\begin{tabular}{lcc|cccc}
\toprule
& \textbf{Ablated} & \textbf{None} & \textbf{SAM2} & \textbf{OneF} & \textbf{BLIP} & \textbf{CLIP}\\
\midrule
\multirow{2}{*}{\rotatebox{90}{\textbf{KI}}} & \textbf{mIoU [\%]} & \textbf{59.4} & 58.4 & 10.3 & 31.5 & 59.3 \\
& \textbf{Cat-mIoU [\%]} & \textbf{69.6} & 68.3 & 19.1 & 51.0 & 69.3 \\
\midrule
\multirow{2}{*}{\rotatebox{90}{\textbf{NU}}} & \textbf{mIoU [\%]} & \textbf{54.9} & 50.2 & 21.4 & 19.6 & 50.5 \\
& \textbf{Cat-mIoU [\%]} & \textbf{62.6} & 59.0 & 26.0 & 28.7 & 60.5 \\
\bottomrule
\end{tabular}
\vspace{-2mm}
\caption{\textbf{Pseudo Labeling Engine} mIoU and category mIoU degradation ablating each engine module, evaluating on Semantic KITTI (KI) and NuScenes (NU).}
\label{tab:pseudo_labels_ablation}
\vspace{-1mm}
\end{table}
\begin{table}[t]
    \centering
    \small
    \begin{subtable}[t]{\linewidth}
      \centering
      \begingroup
      \setlength{\tabcolsep}{10pt}  
      \renewcommand{\arraystretch}{1.0}   
      \begin{tabular*}{0.95\linewidth}{@{\extracolsep{\fill}} l c | c c @{}}
        \toprule
          & \textbf{Full} & \textbf{w/o spline opt.} & \textbf{w/o $f_{iwu}$} \\
        \midrule
          \textbf{mAP [\%]} & \textbf{31.0} & 23.9 & 11.7 \\
          \textbf{NDS [\%]} & \textbf{45.2} & 33.2 & 30.8 \\
        \bottomrule
      \end{tabular*}
      \endgroup
      \caption{Pseudo Bounding Boxes Ablations.}
      \label{tab:spline_opt_abl}
    \end{subtable}
    
    \vspace{0.5em}
    
    \begin{subtable}[t]{\linewidth}
        \centering
        \resizebox{\linewidth}{!}{%
            \begin{tabular}{lcc|ccc}
            \toprule
            & \textbf{Ablated} & \textbf{None} & \textbf{Agorithm \ref{alg:labelrefinement}}
            & \textbf{Accumulation} & \textbf{Occ. Mask} \\
            \midrule
            \multirow{2}{*}{\rotatebox{90}{\textbf{KI}}} & \textbf{mIoU [\%]}      & \textbf{64.9} & 60.7 & 59.4 & 57.0 \\
            & \textbf{cat.-mIoU [\%]} & \textbf{76.2} & 70.4 & 69.6 & 68.5 \\
            \midrule
            \multirow{2}{*}{\rotatebox{90}{\textbf{NU}}} & \textbf{mIoU [\%]}      & \textbf{58.0} & 55.2 & 54.9 & 50.1 \\
            & \textbf{cat.-mIoU [\%]} & \textbf{65.2} & 64.9 & 62.6 & 55.3  \\
            \bottomrule
            \end{tabular}
        }
        \caption{Pseudo Semantic Labels Ablations.}
        \label{tab:abl_occ_mask}
    \end{subtable}

    \vspace{-2mm}
    \caption{\textbf{Ablation Experiments.} Pseudo bounding boxes results (a) ablating the spline optimizer and $f_{iwu}$, and semantic labels results (b) ablating \underline{sequentially} the label propagation algorithm, the accumulation (camera-only) and the occlusion mask in the lifting.}
    \label{tab:combined_ablations}
    \vspace{-1mm}
\end{table}
\section{Conclusion}
We propose an unsupervised pseudo-labeling method that generates semantic labels, bounding boxes, and precise long-range depth from LiDAR, camera and IMU datas recorded in a single driving trajectory. Our approach is based on a geometry-grounded dynamic scene decomposition: we first reconstruct a LiDAR map of the environment, then propagate semantic labels from vision foundation models across each observed point. By detecting and reconciling inconsistencies, we remove moving objects and correct label errors, enabling a truly automatic annotation pipeline that achieves near-oracle performance compared to manual labeling. Our method is not tailored to any specific sensor configuration and generalizes successfully across KITTI, NuScenes and our Long Range datasets. We validate that the generated pseudo-labels achieve state-of-the-art in semantic segmentation and object detection and consistently enhance depth estimation up to $250m$, with improvement of $51.5\%$ in MAE between $80$ and $150$ meters and $22.0\%$ between $150$ and $250$ meters. 
\par
\section{Acknowledgments}
Felix Heide was supported by an NSF CAREER Award (2047359), a Packard Foundation Fellowship, a Sloan Research Fellowship, a Sony Young Faculty Award, a Project X Innovation Award, a Amazon Science Research Award, and a Bosch Research Award.
\par
{
    \small
    \bibliographystyle{ieeenat_fullname}
    \bibliography{main}
}
\end{document}